\documentclass[journal]{IEEEtran}

\usepackage{cite}
\usepackage{amsmath,amssymb,amsfonts}
\usepackage{algorithmic}
\usepackage{graphicx}
\usepackage{textcomp}
\usepackage{amssymb}
\usepackage{pifont}
\usepackage{array}
\usepackage[T1]{fontenc}

\usepackage{hyperref}
\usepackage{pifont}
\newcommand{\xmark}{\ding{55}}
\newcommand{\cmark}{\ding{51}}

\ifCLASSINFOpdf
\else
\fi

\begin{document}

\title{Mammo-Clustering: A Multi-views Tri-level Information Fusion Context Clustering Framework for Localization and Classification in Mammography}

\author{
\thanks{This work is partially supported by grants from the Clinical Research Project of the First Affiliated Hospital of Shenzhen University (2023YJLCYJ019)}

Shilong Yang, Chulong Zhang \IEEEmembership{Member, IEEE}, 
\thanks{Shilong Yang, Chulong Zhang, Xiaokun Liang, and Yaoqin Xie are with the Shenzhen Institutes of Advanced Technology, Chinese Academy of Sciences, 1068 Xueyuan Avenue, Shenzhen University Town, China, 518055.}
Qi Zang, 
\thanks{Qi Zang is with the Qingdao University, Qingdao, China, 266000.}
Juan Yu, Liang Zeng, Xiao Luo, Yexuan Xing, Xin Pan, Qi Li,
\thanks{Juan Yu, Liang Zeng, Xiao Luo, Yexuan Xing, Xin Pan, Qi Li are with the Department of Radiology, The First Affiliated Hospital of Shenzhen University, Health Science Center, Shenzhen Second People's Hospital, 3002 SunGangXi Road, Shenzhen, 518035, China.}
Xiaokun Liang \IEEEmembership{Member, IEEE}, and Yaoqin Xie
\thanks{These authors contributions are equal: Shilong Yang and Chulong zhang}
\thanks{corresponding authors: Xiaokun Liang and Yaoqin Xie (e-mail: xk.liang@siat.ac.cn and yq.xie@siat.ac.cn)}

\thanks{Manuscript received April 19, 2021; revised August 16, 2021.}
}

\markboth{}
{Shell \MakeLowercase{\textit{et al.}}: Bare Demo of IEEEtran.cls for IEEE Journals}

\maketitle

\begin{abstract}
Breast cancer is a significant global health issue, and the diagnosis of breast imaging has always been challenging. Mammography images typically have extremely high resolution, with lesions occupying only a very small area. Down-sampling in neural networks can easily lead to the loss of microcalcifications or subtle structures, making it difficult for traditional neural network architectures to address these issues. To tackle these challenges, we propose a Context Clustering Network with triple information fusion. Firstly, compared to CNNs or transformers, we find that Context clustering methods (1) are more computationally efficient and (2) can more easily associate structural or pathological features, making them suitable for the clinical tasks of mammography. Secondly, we propose a triple information fusion mechanism that integrates global information, feature-based local information, and patch-based local information.
The proposed approach is rigorously evaluated on two public datasets, Vindr-Mammo and CBIS-DDSM, using five independent splits to ensure statistical robustness. Our method achieves an AUC of \(0.828 \pm 0.020\) on Vindr-Mammo and \(0.805 \pm 0.020\) on CBIS-DDSM, outperforming the next best method by 3.1\% and 2.4\%, respectively. These improvements are statistically significant (\(p<0.05\)), underscoring the benefits of Context Clustering Network with triple information fusion. 
Overall, our Context Clustering framework demonstrates strong potential as a scalable and cost-effective solution for large-scale mammography screening, enabling more efficient and accurate breast cancer detection. Access to our method is available at \href{https://github.com/Sohyu1/Tri-Clustering}{https://github.com/Sohyu1/Mammo-Clustering}.
\end{abstract}


\IEEEpeerreviewmaketitle

\section{Introduction}
\IEEEPARstart{R}{ecent} studies highlight that breast cancer, as the most common malignancy among women, has surpassed cardiovascular diseases to become the leading cause of premature mortality among women worldwide \cite{4} \cite{5}. However, breast cancer is also notably amenable to effective prevention and treatment strategies \cite{18}. Early detection of breast cancer is crucial for reducing breast cancer mortality and significantly improving patient prognosis \cite{6} \cite{19}. It will allows for less invasive and more targeted treatment options, thereby reducing the physical and psychological burden on patients \cite{28}.

Mammography is a low-dose, non-invasive X-ray imaging technique \cite{29} that plays a crucial role in the early detection of breast cancer by identifying tumors too small to be palpated, thereby facilitating timely intervention. Several studies have suggested that using mammography for early breast cancer screening can significantly reduce mortality by up to 20\% \cite{1}.

One aspect of the specificity of the mammography issue is that, as a multi-view imaging technique, mammograms are typically acquired from the craniocaudal (CC) and mediolateral oblique (MLO) angles of both the left and right breasts. From a given perspective, the symmetry between the left and right breasts also serves as a critical diagnostic criterion in clinical practice. Consequently, employing a multi-view learning strategy can leverage the complementary information provided by these different imaging angles, thereby enhancing classification performance \cite{63} \cite{65}.
Multi-view learning is a machine learning paradigm that leverages multiple feature sets, or “views,” to improve performance by capturing complementary insights. Widely applied in fields like image analysis, NLP, and bioinformatics, it enhances generalization, robustness to noise, and accuracy by integrating diverse perspectives, often outperforming single-view approaches \cite{41}\cite{42} \cite{44}\cite{45}.
Currently, multi-view learning has become a consensus in the field of mammography analysis.

In addition, mammographic images typically possess extremely high resolution, with lesions occupying only a very small area. And handling extremely high resolution images is a challenge for traditional networks. 
Such as, in Convolutional Neural Networks (CNNs), layer-by-layer downsampling (like pooling) compresses feature maps, potentially leading to the loss of microcalcifications or subtle structures. Accurate localization of these lesions is crucial for diagnosis. 
In the case of Transformers, lesions usually occupy a very small area, and the global attention mechanism of transformers may excessively focus on irrelevant background, introducing noise.
To address these limitations, several methods have been proposed to integrate global and local information to mitigate the loss of local information. 
The GMIC (Globally-Aware Multiple Instance Classifier) introduces an innovative framework that integrates global and local information extracted by ResNet. This concept is akin to multiple instance learning (MIL), where the selection of local information is determined by CAM to identify and extract several local regions of the image \cite{25}. This method dynamically identifies more important regions for patch-based information extraction using a Saliency Map, then segments the original image to obtain patches. It is a weakly supervised approach that requires only image-level labels for patch-level lesion localization. This weakly supervised approach effectively reduces the model’s reliance on high-quality annotated data.

The relevant approach introduces multi-view learning based on GMIC, achieving performance improvement by integrating multi-view feature information through pooling \cite{69}.

\begin{figure*}[ht]
    \centering 
    \includegraphics[width=1\textwidth, angle=0]{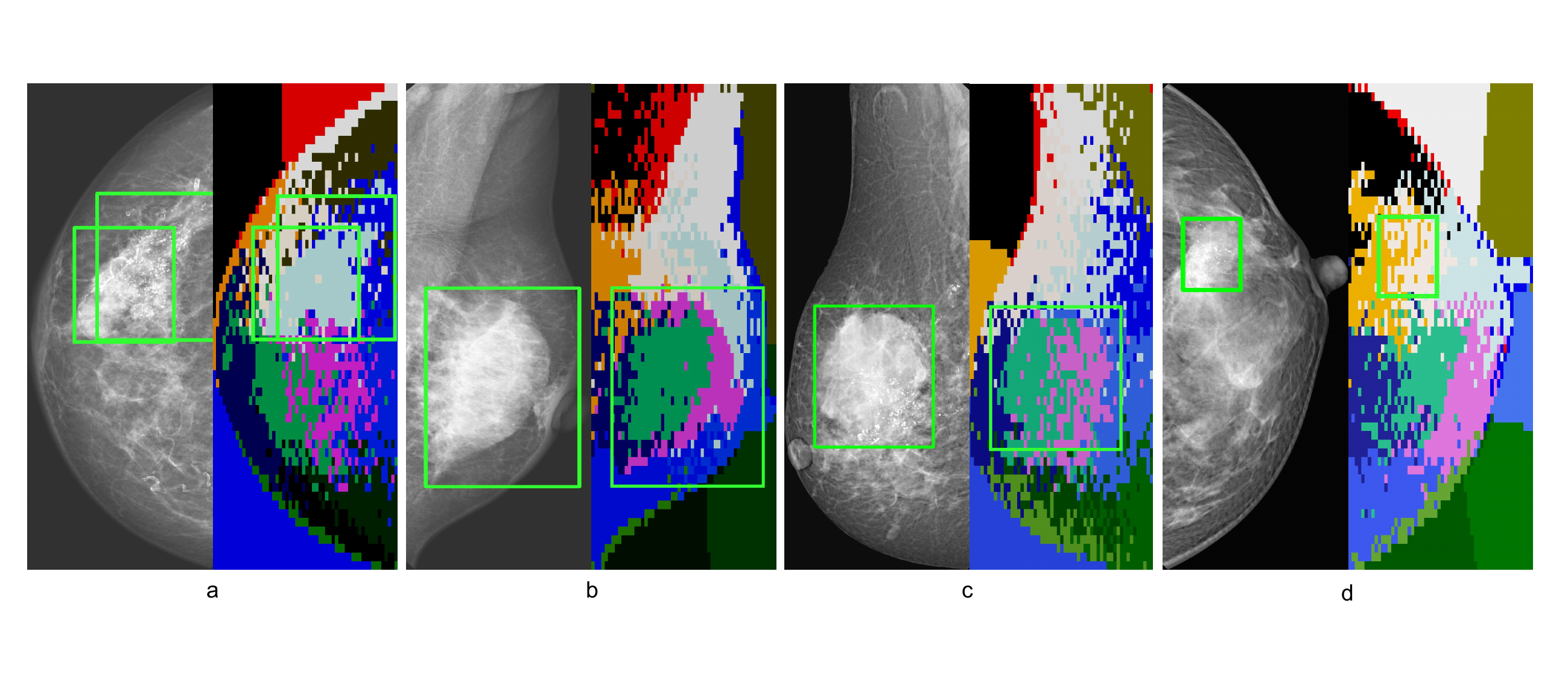}	
    \caption{
    Context Clustering Visualization Diagram. Figures a to d show that the left half of each image shows the original mammogram with annotated suspicious lesions, while the right half presents Contextual clustering visualization, akin to a CNN heatmap and a VIT attention map, with the suspicious lesion locations also outlined. This figure clearly shows that the Context clustering approach effectively identifies and groups suspicious lesion areas in mammography.
    } 
    \label{fig_mom4}
\end{figure*}

These methods are still based on Transformer and CNN, thus the limitations of the traditional paradigms in handling high-resolution images remain unaddressed. Therefore, we opted for the COC network to process mammography images \cite{74}. When using COC (without prior medical image training) for inference, we observed that the network inherently demonstrated a distinct intrinsic clustering capability for breast tissue and lesions, which is visually evident in Figure \ref{fig_mom4}. The clustering results may be more easily correlated with anatomical or pathological features and align with clinical needs.

Furthermore, these methods of feature fusion across different scales have not effectively utilized the information at each scale. For instance, traditional MV methods only employ global information. Conventional MIL methods, typically applied to pathological images, can only extract local information, and in GMIC framework methods, the global information extraction network extracts feature information from the entire image, but the feature information that play a role are actually few (due to the characteristic that mammography lesion areas are sparse), resulting in significant loss in the final pooled feature information, and the feature information from the global network are often wasted. And, in the current model, the local information is patch-based, lacking connections with other parts. It is merely isolated, patch-based information, which inevitably leads to the loss of global characteristics.

\begin{table}[ht]
    \centering
    \caption{Comparison of Model Structures. There A is Multi-view structure, B is with global information, C is with patch-based local information, D is with feature-based local information. }
    \begin{tabular}{l|cccc}
    \hline
    Model & A & B & C & D \\ \hline
    DsMIL\cite{77}          & \xmark & \xmark & \cmark & \xmark \\ \hline
    AbMIL\cite{79}          & \xmark & \xmark & \cmark & \xmark \\ \hline
    TransMIL\cite{76}       & \xmark & \xmark & \cmark & \xmark \\ \hline
    SV Res\cite{63}         & \xmark & \cmark & \xmark & \xmark \\ \hline
    SV SwinT\cite{75}       & \xmark & \cmark & \xmark & \xmark \\ \hline
    GMIC\cite{25}           & \xmark & \cmark & \cmark & \xmark \\ \hline
    MV Res\cite{63}         & \cmark & \cmark & \xmark & \xmark \\ \hline
    MaMVT\cite{65}          & \cmark & \cmark & \xmark & \xmark \\ \hline
    MV GMIC\cite{69}        & \cmark & \cmark & \cmark & \xmark \\ \hline
    Mammo-Clustering(ours)  & \cmark & \cmark & \cmark & \cmark \\ \hline
    \end{tabular}
    \label{table}
\end{table}

To address the limitations of such feature loss, we propose a three-level information fusion architecture, dividing the features when processing mammography images into global information, feature-based local information, and patch-based local information. We believe these three levels of information can maximize the utilization of mammography data. Therefore, we propose a 3-level global-local information fusion mechanism, which can be called the Tri-level Information Fusion Framework (TIFF). We will compared to previous methods (MV, MIL, GMIC), the differences are shown in the table \ref{table}.

We focus on the limitation that the resolution of mammography is too high and where lesions occupy only a small portion of the image, and propose  Weakly Supervised Multi-view Tri-level Information Fusion Context Clustering Network. Our primary contributions include:
\begin{itemize}
    \item[$\bullet$] This work introduces a novel non-CNN, non-attention-based feature extraction method using \emph{Context clustering} for early breast cancer screening in mammography.
    \item[$\bullet$] We propose a fusion mechanism named Tri-level Information Fusion Framework (TIFF) that integrates global, feature-based local, and patch-based local information, with enhanced focus on local details.
    \item[$\bullet$] Our method achieves state-of-the-art accuracy with the lowest parameter count among comparable techniques, ensuring efficiency.
\end{itemize}

\section{Method}
Our research aims to design a rapid and reliable system for early breast cancer screening, with the dual objectives of reducing the workload on medical professionals and expanding access to screening opportunities for women in underdeveloped regions.

This section details the architecture of our proposed weakly supervised multi-view network. In Subsection 1, we summarize the overall workflow of our method for early breast cancer screening in mammography. Subsection 2 provides a comprehensive explanation of the key components of our framework, highlighting their individual contributions to the system's efficacy. Finally, in Subsection 3, we introduce the loss functions utilized during training, elaborating on their roles in optimizing the network's performance and ensuring robust learning.

\subsection{Overall Framework}
\begin{figure*}[ht]
    \centering
    \includegraphics[width=1\textwidth, angle=0]{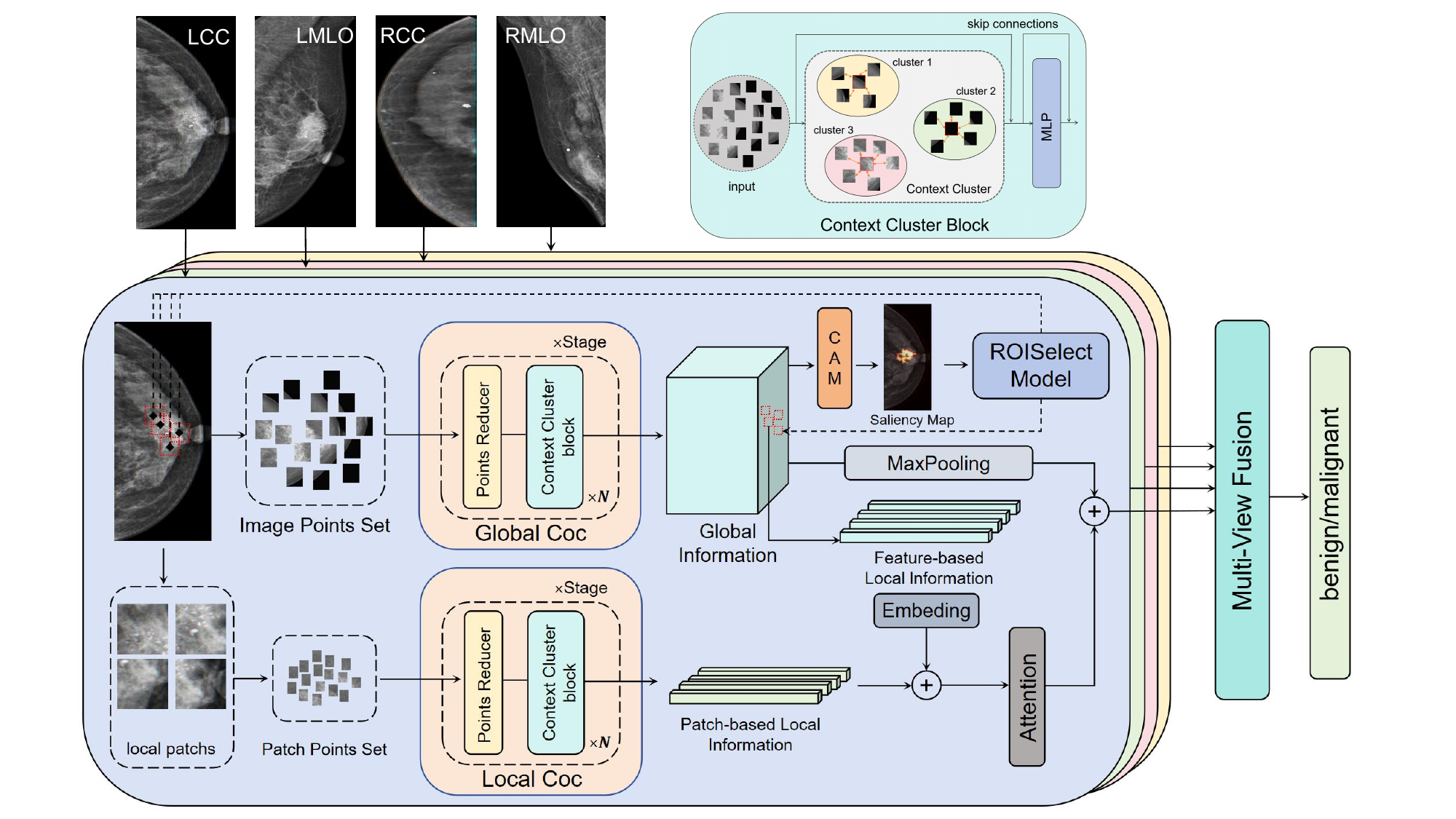}	
    \caption{
    Architecture of the proposed model. Images from four perspectives are enhanced into point sets and processed via a multi-level Context Clustering module, Global Coc, to extract global information. This module includes point reducers and context cluster blocks. The ROISelectModel utilizes this global information to select patch-based images, which are processed through another Context Clustering module, Local Coc, to generate patch-based local information. This is fused with feature-based local information derived from the global information to produce local information. Subsequently, local and global information are combined to create single-view fusion information. Fusion information from each perspective is integrated across views and regressed to produce the final output.
    } 
    \label{fig_0}%
\end{figure*}

The proposed model for mammography classification can be formulated as follows:

For each image $I$ in the given view, we enhance all points into 5-dimensional information points containing color and position data to obtain the set of points $S \in \mathbb{R}^{5, w \times h}$, where $w \times h$ is equal with the number of points.

\subsubsection{Global Information Extraction}
\label{sec:Global Information Extraction}
The set $S$ is input to the first Context-Clustering network to obtain global information $F_g$:
\begin{equation}
    F_{g} = f_{global}(S)
\end{equation}

where $f_{global}$ represents the Context-Clustering network for global information extraction.

The Saliency Map is obtained by processing the feature information $F_{g}$ extracted by global network through a feedforward network, denoted as $I_{map}$.

And based on the Saliency map, the ROI selection module outputs a set of location information $P$. This selection is independent of dataset annotations, and the positions of the selected \text{n} patches can be expressed as:
\[
P = \{ p_{1}, p_{2}, \ldots, p_{n}\}
\]

where $p_{n}$ is a coordinate representing a position, written as $(x_{n}, y_{n})$. The value of \text{n} is manually set.

With $P$, we can extract \text{n} patches $\tilde{I}$ from the original image $I$ and extract \text{n} feature-based local information $F^{n}_{fl}$ from the global information $F_{g}$. For the specified number of patches \text{n}, we recommend setting it to 4. This is because when the number of suspicious lesion locations exceeds this value, some lesions may be missed, thereby affecting the screening performance of the model. However, when the number of patches is greater than the number of suspicious lesion locations, the attention mechanism dynamically adjusts the weights of these patches, allowing the model to focus more on patches that truly contain lesion locations. Therefore, in this case, the size does not impact the model’s performance.

\subsubsection{Patch-based Local Information Extraction}
Each selected patch $\tilde{I}_{i}$ is treated as a new image, re-enhanced based on each point to obtain its five dimensional point set $\tilde{S}_{i}$. And processed through a second Context-Clustering network to obtain patch-based local information $F_{pl}^{i}$, where $i\in \{1, 2, ..., n\}$.

\subsubsection{Information Fusion and Attention Mechanism}
The local information $F_{l}$ from all patches is generated by fusing all feature-based local information and all patch-based local information:
\begin{equation}
    F_l = f_{atten}(F_{fl}\oplus F_{pl})=f_{atten}(\sum_{i=0}^{n}F_g(x_i, y_i) \oplus \sum_{i=0}^{n}f_{local}(\tilde{S}_{i}))
\end{equation}
The operation $\oplus$ represents the fusion operate of the two types of information, after that processed through an attention mechanism to enhance relevant features:

Then, the local information $F_l$ is fused with the original global information $F_g$, resulting in multi-instance fusion information $F_f$ from single-view.

\subsubsection{View Fusion and Classification}
Process the images $I_{view}$ from the four views (bilateral craniocaudal (CC) and mediolateral oblique (MLO)) using the aforementioned procedure to obtain single-view fusion information. This information is then integrated for multi-view fusion, which is used for the final binary classification, resulting in the early screening model’s output:
\begin{equation}
    F_{fusion} = {f}_{fuse}(F_f^{\text{lcc}}, F_f^{\text{lmlo}}, F_f^{\text{rcc}}, F_f^{\text{rmlo}})
\end{equation}

where $F_f^{\text{lcc}}$ represents the fusion feature of LCC images, other similar situations. ${f}_{fuse}$ is a fusion structure. In addition to integrating features $F_{fusion}$, ${f}_{fuse}$ also merges $F_{g}$ and $F_{l}$ from different views.

Finally, $F_{fusion}$ performs the final classification.

This formulation encapsulates the entire process of the weakly supervised multi-instance multi-view network for mammography classification.

\subsection{Detailed Network Structure}
\subsubsection{From Image to Set of Points}
The scale of an image can be expressed as (3, h, w), where 3 represents the RGB channels, and h and w are its height and width. We enhance each pixel by considering it as a 5-dimensional data point containing color and positional information (r, g, b, x, y). After this enhancement, the image can be represented as a set of $h \times w$ 5-dimensional data points, with a scale of $(h \times w,  5)$ \cite{74} . We can then perform feature extraction through simple clustering. From a global perspective, the image is viewed as a collection of unordered discrete data points with color and positional information. Through clustering, all points are grouped into clusters, each containing a centroid. Since each point in the set includes color and positional information, this clustering implicitly incorporates spatial and image information. 

\subsubsection{Context Clustering}

Context Clustering performs deep feature extraction through a hierarchical structure of Context Clustering Blocks, similar to convolutional networks \cite{74}. A point reducer is placed before each block to reduce the number of points, thereby enhancing computational efficiency. Subsequently, aggregated features are adaptively assigned to each uniformly selected anchor point within clusters based on similarity, and neighboring points are connected and fused through linear projection.

\begin{figure}
	\centering 
	\includegraphics[width=0.5\textwidth, angle=0]{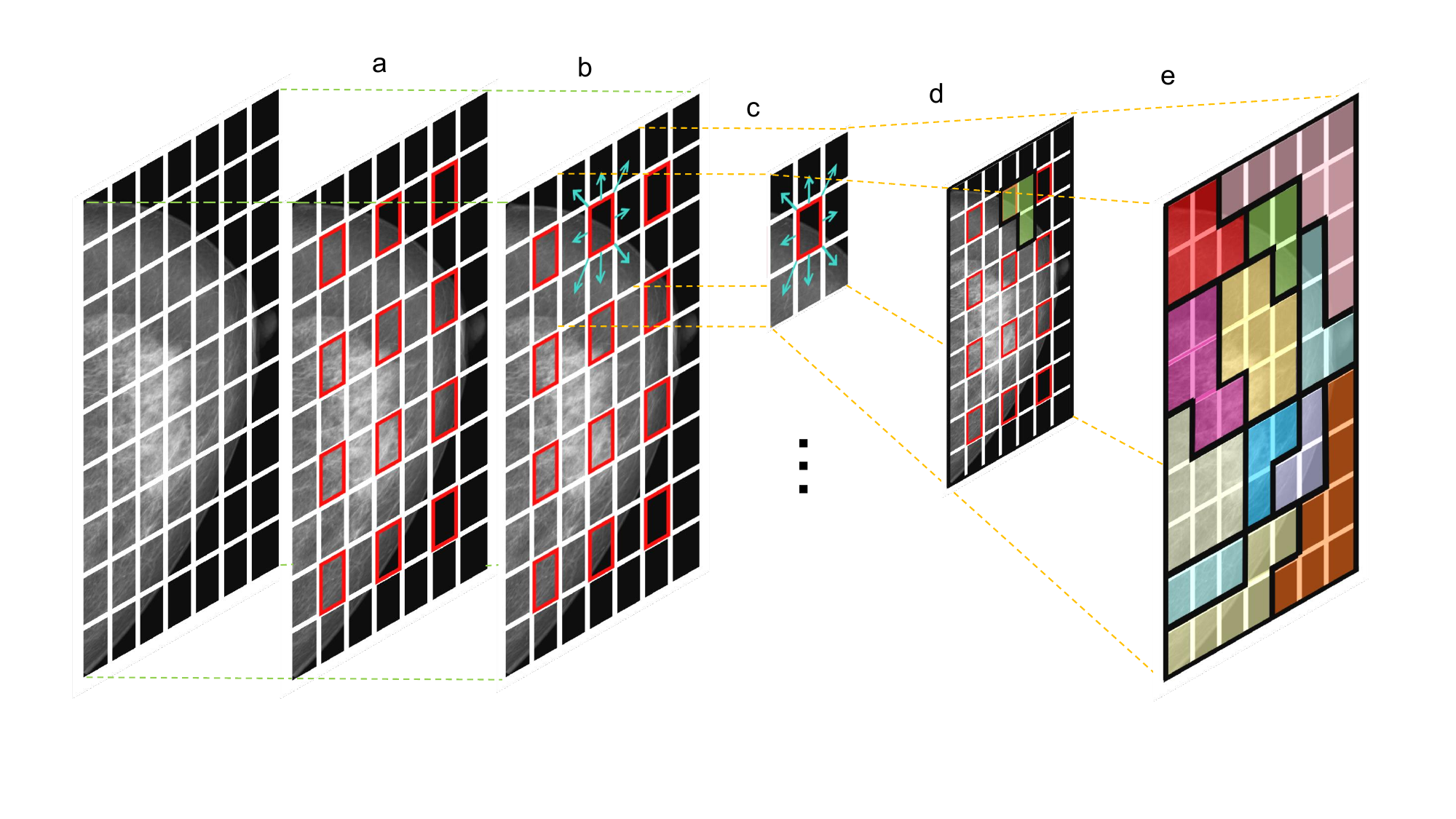}	
	\caption{
        A visual explanation of Context-Clustering. This clustering consists of five components: selecting central anchor points, identifying neighbors for each anchor, calculating features for each anchor, performing similarity analysis based on these anchors, and representing all clusters on the chart.
        }
	\label{fig_mom1}
\end{figure}

As illustrated in Figure \ref{fig_mom1}, the input image undergoes point set transformation, and then, in step \textbf{a}, $n$ central anchor points are uniformly selected in the space. the method is similar to those in SuperPixel \cite{73} and SLIC \cite{66}.

The selected central anchor points are highlighted with red boxes in the figure. In step \textbf{b}, for each central anchor point, $k$ neighbors are identified, indicated by arrows in the figure. The value of $k$ can be 4 or 8, as determined manually, and it can also be the four neighbors in the up, down, left, and right directions, in which case $k$ equals 4. 

Step \textbf{c} involves calculating the features of a central anchor point determined by itself and its $k$ neighboring points, illustrated in the figure for the case where the number of neighbors is 8. The calculation process is: 
\[
P^{x} = \frac{\left( P^{x} + \sum_{q=1}^{k} p_q^x \right)}{k+1} 
\]
where $P^{x}$ denotes the feature of the central anchor point in the $x$ dimension, $x \in \{r,g,b,h,w\}$. Meanwhile, $p_q^x$ represents the feature of the $q$-th neighboring point in the $x$ dimension, $q \in \{0, 1, 2, \ldots, k\}$.

After computing the features for all central anchor points, a similarity analysis is conducted between all points in the point set and each central anchor point’s features in step \textbf{d}. Each point is assigned to the cluster of the central anchor point with which it has the highest similarity. The steps for the similarity analysis is conducted by computing the pairwise cosine similarity matrix $M$ between a point and set of central points: 
\[
M(P_i, P'_j) = \frac{P_i \cdot P'_j}{|P_i| |P'_j|}
\]
where $M(P_i, P'_j)$ is pairwise cosine similarity matrix. $P_i$ is $i$-th central anchor point, $i \in \{0, 1, 2, \ldots, n\}$. $P'_j$ is $j$-th point in the image point set. 

Finally, in step \textbf{e}, all clusters are combined, resulting in the desired clustering outcome for the entire image.

Since each point in the set includes color and position information, this grouping implicitly incorporates spatial and image information \cite{74}.

\subsubsection{CAM}
Feature visualization is computed by a feedforward network to generate saliency maps. This CAM module consists of convolution layers with a kernel size of 1 and is integrated into the Global Network to enable iterative optimization.

\subsubsection{ROI Selection Module}
The saliency map is normalized and divided into regions with a height of $h_{crop}$ and a width of $w_{crop}$ regions for greedy ROI search, $h_{crop}$ and $w_{crop}$ are dynamically adjusted based on the original image and saliency map scales. In each iteration, the algorithm greedily identifies each region and selects the one with the largest total weight, as determined by average pooling, among all current regions. The coordinates of this region are added to a list, and a mask flag is applied to the region to prevent redundant selection. The coordinates of these regions will be mapped to the size of the original image to obtain patch-based images.

\begin{figure*}[ht]
	\centering 
	\includegraphics[width=1\textwidth, angle=0]{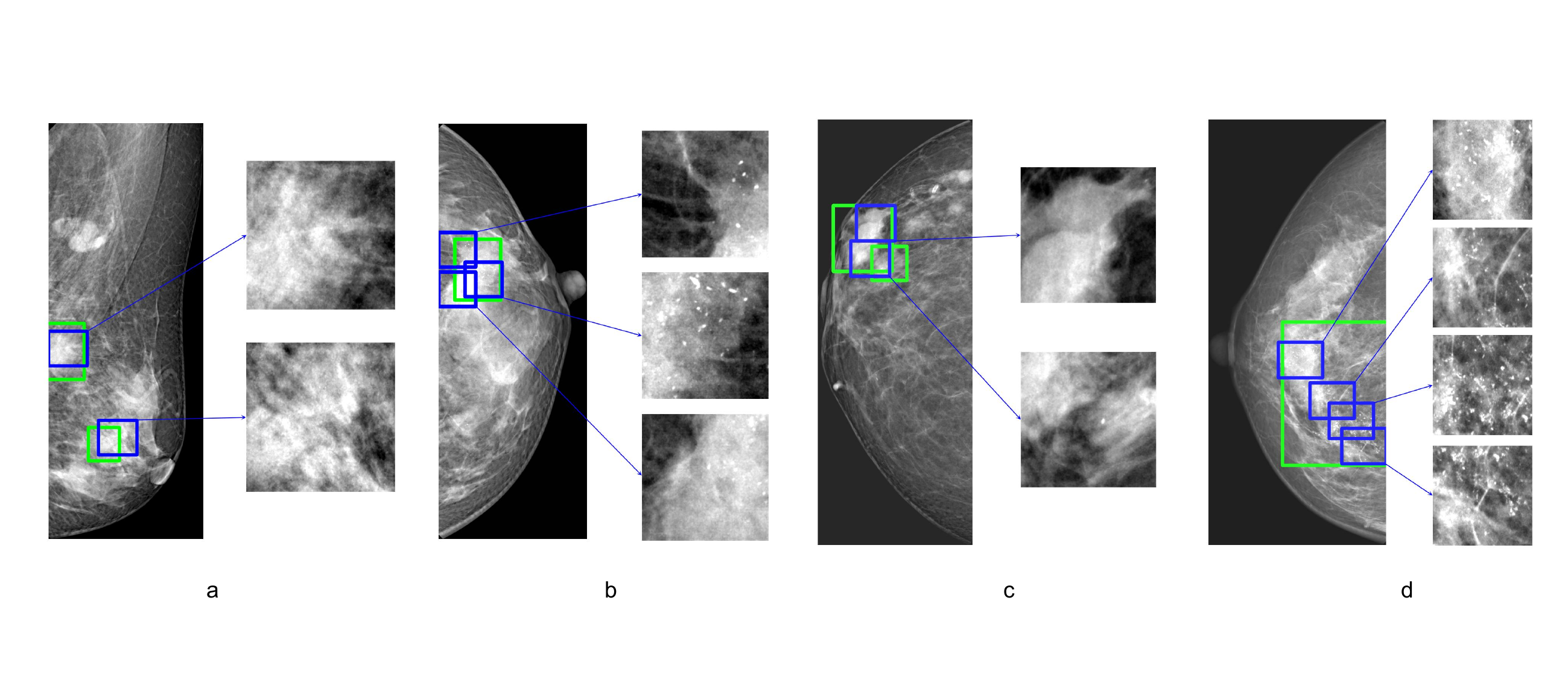}	
	\caption{
Visualization of patch-based images extracted by the model. The green box on the mammography indicates the location of the suspicious lesion, while the blue box represents the patch-based images selected by the model. We can observe that the model’s extracted patch-based images perform exceptionally well, and the magnified images clearly show calcifications and masses.
} 
	\label{fig_mom2}%
\end{figure*}

Figure \ref{fig_mom2} visualizes the patch-based images selected by the ROISelectModule, along with the patches’ positions on the source image and their comparison with the locations of suspicious lesions.
\begin{table*}[htp]
\centering
\caption{Performance of each model on two datasets. SV and MV represent Single-View and Multi-Views, respectively. Backbone of the network after "-".}
\begin{tabular}{lccccccc}
 \hline
 & \multicolumn{3}{c}{\textbf{Vindr-Mammo}} & \multicolumn{3}{c}{\textbf{CBIS-DDSM}} \\
 \hline
 &AUC & ACC & F1 score &AUC & ACC & F1 score & Params \\
 \hline
DsMIL\cite{77}               & $0.605 \pm 0.02$ & 0.730 & 0.781 & $0.697 \pm 0.02$ & 0.500 & 0.583 & \textbf{202116} \\
AbMIL\cite{77}               & $0.618 \pm 0.02$ & 0.776 & 0.825 & $0.726 \pm 0.02$ & 0.571 & 0.671 & 396547   \\
TransMIL\cite{76}            & $0.631 \pm 0.02$ & 0.888 & 0.890 & $0.739 \pm 0.02$ & 0.635 & 0.637 & 2544402  \\
SV Res18 \cite{63}           & $0.727 \pm 0.02$ & 0.783 & 0.821 & $0.719 \pm 0.02$ & 0.646 & 0.639 & 1477025  \\
SV SwinT\cite{75}            & $0.731 \pm 0.02$ & 0.651 & 0.719 & $0.724 \pm 0.02$ & 0.651 & 0.599 & 14184625 \\
GMIC-Res18 \cite{25}         & $0.793 \pm 0.02$ & 0.847 & 0.878 & $0.778 \pm 0.02$ & 0.682 & 0.680 & 22487298 \\
MV Res18 \cite{63}           & $0.740 \pm 0.02$ & 0.753 & 0.796 & $0.731 \pm 0.02$ & 0.676 & 0.641 & 6128546  \\
MaMVT \cite{65}              & $0.770 \pm 0.02$ & 0.882 & 0.867 & $0.749 \pm 0.02$ & 0.649 & 0.649 & 30730082 \\
MV GMIC-Res18 \cite{69}      & $0.797 \pm 0.02$ & 0.887 & 0.879 & $0.781 \pm 0.02$ & 0.699 & 0.691 & 22686871 \\
MV GMIC-SwinT                & $0.799 \pm 0.02$ & 0.874 & 0.854 & $0.785 \pm 0.02$ & 0.694 & 0.694 & 28873234 \\
Mammo-Clustering(ours)       & $\textbf{0.828} \pm 0.02$ & \textbf{0.919} & \textbf{0.906} & $\textbf{0.805} \pm 0.02$ & \textbf{0.709} & \textbf{0.709} & 9805459 \\ \hline
\end{tabular}
\label{Table2}
\end{table*}

\subsubsection{Attention Module}
In our model, we employ attention mechanisms in two locations to integrate multi-instance and multi-view information, respectively.

\textbf{Approximate Multi-Instances information fusion:} Attention is employed here to mitigate the impact of redundant image patches on training. After the global network, the ROI select module to choose $n$ patch-based images, a number set manually. This implies not all patch-based images carry beneficial information, and some may be redundant. Considering and integrating all the information from these patch-based images could significantly impair our network. Therefore, an attention module is added before integrating local and global information, allowing the model to learn how to filter out irrelevant local information.

The attention mechanism receives feature representation of patch-based images $F_{l}$, shaped as $(batchsize, k, dim)$, where $batchsize$ and $k$ are manually set parameters; the former defines the batch size during training, while the latter specifies the number of patches required for multiple instance learning. The size of $dim$ varies depending on the model.

Firstly, we use a neural network layer with a simple linear transformation $f_{weights}$ and a softmax function to compute attention weights. Subsequently, the attention weights are multiplied pointwise with the feature representation of the block-based image $F_{l}$ to obtain the final implicit representation $F_{a}$.

\[
F_{a} = F_l \odot \text{softmax}(f_{weights}(F_{l}))
\]
where $\odot$ represents the stationary point multiplication algorithm.

\textbf{Multi-view information fusion: }
Attention is employed here to mitigate the impact of redundant image views on training.

In multi-view learning, not all information from each view is necessarily classified as malignant. However, If a single view exhibits malignant characteristics, the instance should be classified as malignant. Therefore, we introduce an attention mechanism to enable the model to autonomously filter out irrelevant view information, enhancing classification accuracy.

The attention mechanism processing is largely consistent with multi-instance fusion attention. However, in multi-view attention, this attention module processes not only the $F_{a}$ fused by the multi-instance attention module but also $F_{g}$ and $F_{l}$. This is because we will perform targeted optimization on different network structures based on the losses obtained from various features.

\subsubsection{Embedding Module}
\label{Embedding module}
The embedding module in the model primarily aligns feature-based local information $F_{fl}$ with patch-based local information $F_{pl}$ before their integration. Here, we employ a trainable MLP to align the scales. We designed relevant ablation experiments to verify its effectiveness, the effectiveness can be observed in Table \ref{Table4}.

\subsubsection{Maxpooling Module}
We employ max-pooling to fold and align global information $F_{g}$, facilitating better integration with local information fused through the attention module.

\subsection{Loss Function}
We chose a composite loss function to achieve targeted optimization of different components.
\[
L = \alpha \cdot L_{global} + \beta \cdot L_{local} + \gamma \cdot L_{fusion} + \delta \cdot L_{map}
\]

After the multi-view fusion module, we retain not only the fused information for regression but also intermediate features such as global information, local information, and saliency maps. These features are used to compute a composite loss function for precise optimization of each part of the network. And we determine the sensitivity of the loss function to different types of data through component analysis.

$L_{global}$ is calculated using the global information obtained from multi-view fusion and the ground truth values. The loss function chosen here is BCELoss. And $L_{map}$ will be calculated from the saliency-map, it is the weighted average intensity of the saliency-map under the L1 norm. The $L_{global}$, combined with $L_{map}$, indicates the quality of the Global Network and further to determining the adjustment magnitude for the Global Network to enhance the accuracy of locating patch-based images. BCEWithLogitsLoss function is used for both $L_{local}$ and $L_{fusion}$. $L_{local}$ represents the quality of the local network, calculated from local information and ground truth values, determining the adjustment magnitude for the Local Network to enhance the feature extraction capability of the Local Network. $L_{fusion}$ represents the model’s final classification error, driving the overall model training. The weights $\alpha$, $\beta$, $\gamma$, and $\delta$ represent the proportion of each loss, all manually set, Here, we consider $L_{global}$, $L_{local}$, and $L_{fusion}$ to have equal importance, so we recommend setting $\alpha$, $\beta$, and $\gamma$ uniformly to 1. Meanwhile, $L_{map}$ represents the weighted average intensity of the saliency map calculated under the L1 norm, resulting in a relatively large value. Therefore, we suggest setting $\delta$ to a value less than 0.001.

\section{Experiment and Result}
\subsection{Datasets}
\subsubsection{Vindr-Mammo}
The Vindr-Mammo \cite{67} dataset is a large-scale, annotated collection of digital mammographic images aimed at advancing breast cancer detection and diagnosis through machine learning. It includes thousands of images sourced from diverse populations, with detailed annotations such as lesion types, BI-RADS categories, and precise lesion locations. This dataset is designed to support the development of robust AI models by providing a wide variety of cases, including both normal and abnormal findings, thus enhancing the generalizability and accuracy of diagnostic algorithms. 
\subsubsection{CBIS-DDSM}
The CBIS-DDSM (Curated Breast Imaging Subset of the Digital Database for Screening Mammography) \cite{68} dataset is a widely used resource in the field of breast cancer research. It comprises digitized film mammograms, which have been meticulously annotated with information such as lesion boundaries, types (e.g., calcifications, masses), and pathology-confirmed labels (benign or malignant). The dataset also includes patient metadata and additional clinical information, making it an invaluable tool for training and validating computer-aided detection and diagnosis systems. Its comprehensive nature and established use in the research community make it a benchmark for evaluating the performance of mammography-based AI models.

Both Vindr and CBIS-DDSM provide detailed annotations of lesion locations, but such annotation tasks are generally high-cost. By employing weakly supervised learning to enable the network to autonomously localize lesion positions, the cost of dataset creation can be significantly reduced.

\begin{table}[ht]
\caption{The composition of data for the two datasets.}
\begin{tabular}{l c c c c c c c} 
\hline
 & \multicolumn{3}{c}{\textbf{Vindr-Mammo}} & & \multicolumn{3}{c}{\textbf{CBIS-DDSM}} \\
 \hline
    & benign  & malignant & total && benign  & malignant & total\\ 
 \hline
 train & 3614 & 385 & 3999 && 629 & 660 & 1289\\
 test & 904 & 96 & 1000 && 185 & 146 & 331 \\
 total & 4518 & 481 & 4999 && 814 & 806 & 1620\\
 \hline
\end{tabular}
\label{Table1}
\end{table}

\subsection{Evaluating Indicator}
In breast cancer early screening models, several evaluation metrics are commonly used to assess the performance of the classification models. Here are the definitions and significance of each metric along with their respective formulas:
\subsubsection{Classification indicators}
\begin{itemize}
    \item AUC (Area Under the Curve): AUC means the area under the receiver operating characteristic (ROC) curve. The ROC curve uses the true positive rate for mammography benign-malignant classification as the y-axis and the false positive rate as the x-axis. It provides an aggregate measure of performance across all possible classification thresholds. A higher AUC value indicates a better model performance, with 1 representing a perfect model and 0.5 a random guess.
    \[ 
        \mathrm{AUC}=\int_{0}^{1} T P R \,\,\,dF P R
    \] 
    where $TPR$ is the true positive rate, and $FPR$ is the false positive rate.
    
    \item ACC (Accuracy): Accuracy is the proportion of true results (both true positives and true negatives) among the total number of cases examined, the corresponding clinical term is “specificity”. It gives a straightforward measure of how often the classifier is correct.
    \[ 
        \text{ACC} = \frac{TP + TN}{TP + TN + FP + FN}
    \] where 
    $TP$, $TN$, $FP$ and $FN$ represent the numbers of true positives, true negatives, false positives, and false negatives, respectively.

    \item F1 Score: The F1 Score is the weighted average of Precision and Recall. This score takes both false positives and false negatives into account. Given the long-tail issue in the data, we selected the micro F1 score as the evaluation metric, it is particularly useful when the class distribution is uneven. 
    \[\text { F1 Score } = 2 \cdot \frac{ Precision \cdot Recall}{Precisionn + Recall} \]
    
\end{itemize}

\subsubsection{Localization indicators}
\begin{itemize}
    \item MDR(Miss Detection Rate):
    MDR is defined as the percentage of the number of undetected suspicious lesion areas $N_{miss}$ relative to the total number of suspicious lesion areas $N_{gt}$. Because, in clinical practice, we are more concerned about lesions being undetected, i.e., false negatives, rather than false positives.
    \[
    \textit{MDR} = \frac{N_{miss}}{N_{gt}}
    \]

    \item Recall:
    Recall is a metric used in object detection to evaluate a model’s ability to identify all relevant objects in an image. It measures the proportion of actual positive instances (i.e., objects that should be detected) correctly identified by the model. In this context, it reflects the model’s capability to detect all existing lesions.
    \[
    Recall = \frac{TP}{TP + FN} 
    \]
\end{itemize}

These metrics collectively provide a comprehensive evaluation of the performance of breast cancer screening models, helping to understand their strengths and weaknesses in various aspects of classification.

\subsection{Comparative Experiment}

In this study, we evaluated several models on two datasets: Vindr-Mammo and cbis-ddsm. The performance metrics considered were AUC, ACC, and F1 score and so on.

\subsubsection{Assessment of classfication}
For the Vindr-mammo, ours model achieved the highest performance across all metrics, with an AUC of $0.828 \pm 0.02$, ACC of $0.919$, and F1 score of $0.906$. For Single-View task models, the GMIC-Res18 with a Global-local structure achieves an AUC of 0.793, significantly outperforming other Single-View model, which SV Res18 and SV SwinT had AUCs of $0.727 \pm 0.02$ and $0.731 \pm 0.02$, respectively. For Multi-views task models, our model achieves an AUC of 0.828, significantly surpassing other models. Multi-views GMIC-SwinT also showed competitive performance with an AUC of $0.799 \pm 0.02$, but its ACC and F1 score were lower at $0.874$ and $0.854$. 

On the CBIS-DDSM dataset, the ours model again demonstrated superior performance with an AUC of $0.805 \pm 0.02$, ACC of $0.709$, and F1 score of $0.709$, obtained the highest performance across on this dataset. Multi-View GMIC-Res18, which had an AUC of $0.781 \pm 0.02$ and an F1 score of $0.699$. 
Noticeably, the Single-View MIL method shows promising potential on this dataset, with TransMIL achieving an AUC of $0.739 \pm 0.02$. This patch-based information approach significantly outperforms methods that focus solely on global information.

In this this two datasets, the advantages of Context Clustering and Tri-level Information Fusion architecture are more pronounced than other information fusion method, showing significant advantages in AUC.

\subsubsection{Model Complexity}
In terms of model complexity, measured by the number of parameters, the ours model had 98.05 million parameters. Other smaller networks often cannot achieve the accuracy of our model and show a significant gap. This is efficient compared to the MaMVT with 30.73 million parameters and the Multi-View GMIC with 22.68 million parameters, considering the performance gains achieved.

\subsubsection{ROC curve}

\begin{figure*}[ht]
\centering 	
\includegraphics[width=1\textwidth, angle=0]{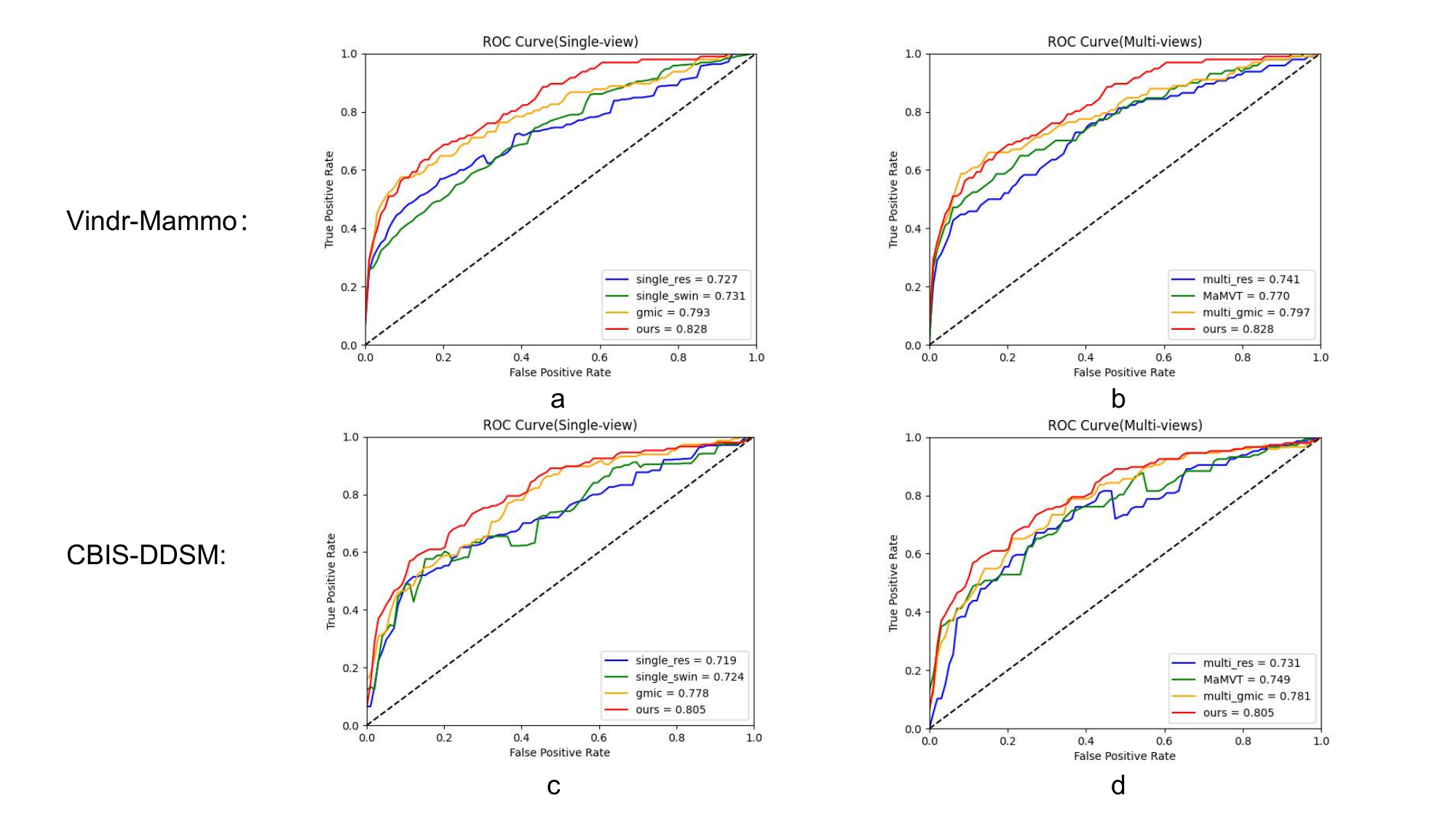}	
	\caption{
    Comparison of ROC curves of different models on two public datasets. Figures a and b compare the ROC curves of our model with other Single-view and Multi-view architectures on the Vindr-mammo dataset. Figures c and d present the ROC curves comparison on the CBIS-DDSM dataset.
} 
	\label{fig_mom3}
\end{figure*}

The ROC curve in figure \ref{fig_mom3} provides insights that cannot be obtained from tables alone. 
Analyzing the ROC curve, we observe that most models, except ours, exhibit a concave shape in the middle. This is due to class imbalance in the data, further validating the effectiveness of our model’s architecture.

Overall, our model offers a robust and efficient approach, achieving state-of-the-art performance on both datasets, surpassing the second-best AUC by over 0.02, with fewer parameters. The global-local architecture proves effective for both multi-view and single-view models. Additionally, the multi-view learning approach enhances model performance.

\subsubsection{Assessment of localization}

We compared the performance of three weakly supervised lesion localization models without comparing them to fully supervised models. Our aim was to validate the feasibility of training weakly supervised models on datasets without annotated lesion regions, which would significantly reduce the burden of creating mammography datasets.

\begin{table}[!ht]
    \centering
    \caption{
    Comparison of Different Weakly Supervised Models in Lesion Localization Tasks.
    }
    \begin{tabular}{l c c}
    \hline
        ~             & MDR   & Recall \\ \hline
        GMIC-Res18    & 0.433 & 0.476  \\ 
        MV GMIC-Res18 & 0.336 & 0.643  \\ 
        MV GMIC-Res34 & 0.412 & 0.479  \\ 
        MV GMIC-SwinT & 0.322 & 0.650  \\ 
        \textbf{Ours} & \textbf{0.294} & \textbf{0.685} \\ \hline
    \end{tabular}
    \label{Table5}
\end{table}

From Table \ref{Table5}, we observe that our model achieves the lowest missed detection rate (MDR) of 0.294 and the lowest recall rate of 0.476, demonstrating its potential. All the comparative models employ the same weakly supervised lesion localization approach as our model, enabling lesion location identification using only classification labels, making them suitable for comparison. We selected four different models encompassing both CNN and attention mechanism paradigms and conducted evaluation analysis on two metrics: MDR and recall. Compared to the second-best model, our approach shows an improvement of approximately 0.03 in both MDR and recall.

\subsubsection{Visual comparison}
\begin{figure*}[ht]
    \centering 
    \includegraphics[width=1\textwidth, angle=0]{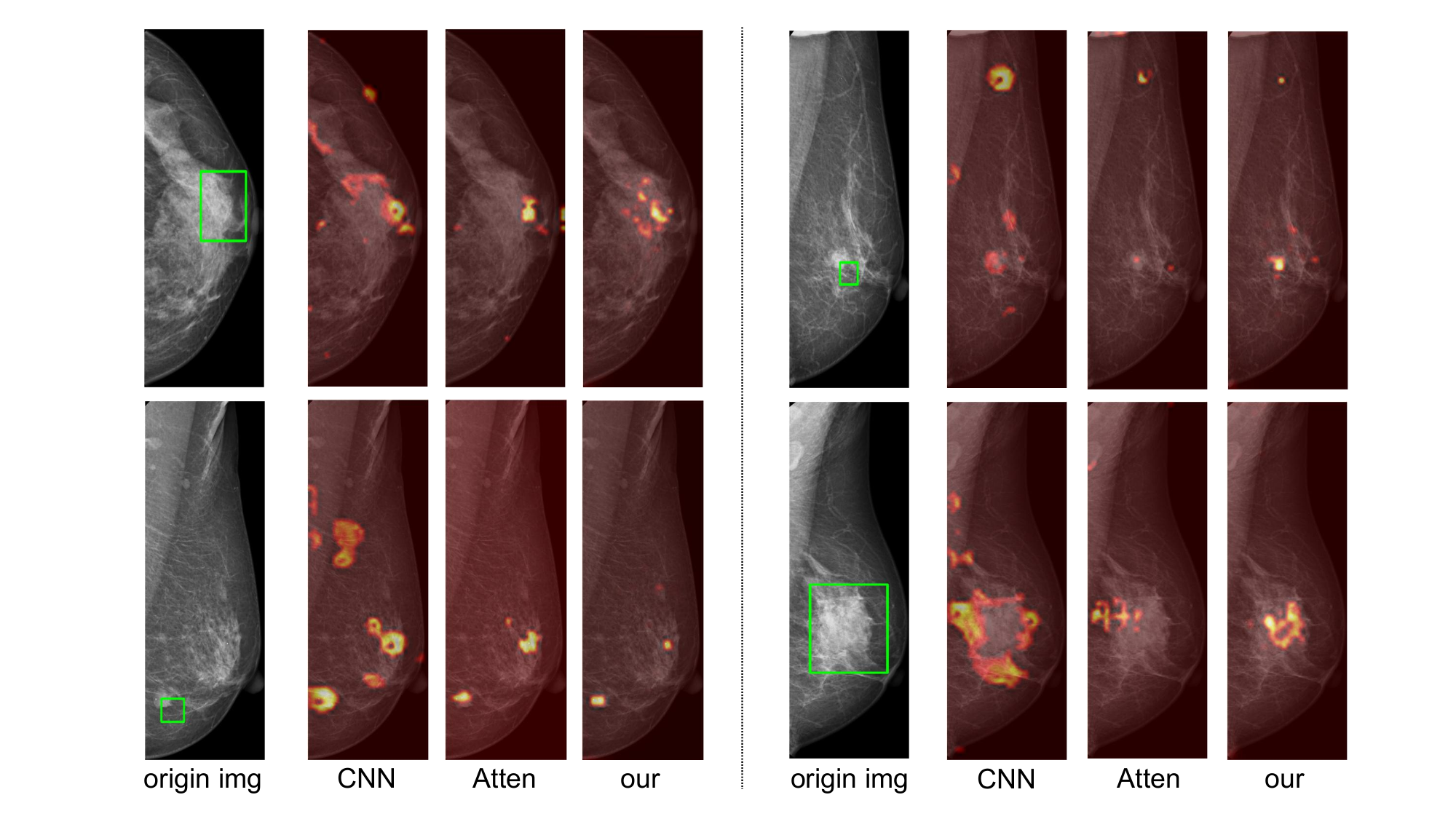}	
    \caption{Comparison of activation map visualizations from different feature extraction paradigms. The green boxes on the original images indicate the annotated true locations of the lesions.
    } 
    \label{cam}
\end{figure*}

We visualized the feature activation maps of different feature paradigms, as shown in Figure \ref{cam}. In this comparison, we selected representative models, ResNet and ViT, as exemplars of CNN and attention mechanisms, respectively, to compare with the Context Clustering paradigm. 

It is evident that the activation maps based on CNN perform the worst. although they sometimes identify the locations of regions of interest, they exhibit low contrast. Activation maps based on attention mechanisms and clustering both achieve relatively clear and accurate localization of regions of interest. However, the activation maps based on Context Clustering are cleaner, indicating they are subject to less interference.

\subsection{Ablation Experiment}

\subsubsection{Different Feature Extraction Paradigm}

This ablation study aims to demonstrate the superiority of Context Clustering in feature extraction performance for Mammography through numerical analysis.

\begin{table}[ht]
\centering
\caption{
Performance of different feature extraction paradigm on the Vindr-Mammo
}
\begin{tabular}{l c c c} 
 \hline
                     & AUC  & ACC & F1 score \\ 
 \hline
 SV Res18(CNN-base) \cite{63}        & $0.727 \pm 0.02$ & 0.783 & 0.821 \\
 SV SwinT(Atten-base) \cite{75}      & $0.731 \pm 0.02$ & 0.651 & 0.719 \\
 SV Coc(Clustering-base) \cite{74}   & $\textbf{0.762} \pm 0.02$ & \textbf{0.794} & \textbf{0.833} \\
 \hline
\end{tabular}
\label{Table3}
\end{table}

The table clearly demonstrates the superiority of the Context Clustering architecture, achieving the highest AUC as well as the best ACC and F1 scores in single-view learning, indicating its balance in mammography tasks. $SV$ represents a single-view learning approach.

\subsubsection{Different Information Fusion Method}

We identified two distinct sources of local information: patch-based local information and feature-based local information. Moreover, this feature-based local information has been overlooked in existing work.

The aim of this ablation study is to validate the effectiveness of our Tri level Information Fusion mechanism that combines these two types of local information with global information.

\begin{table}[ht]
\centering
\caption{
Performance of different information fusion method on the Vindr-Mammo
}
\begin{tabular}{l c c c} 
 \hline
    & AUC  & ACC & F1 score \\ 
 \hline
 $\textit{Global}$   & $0.783 \pm 0.02$ & 0.815 & 0.852 \\
 $\textit{Global + Patch-based local}$ & $0.810 \pm 0.02$ &0.890& 0.891 \\
 $\textit{Global + Feature-based local}$ & $0.806 \pm 0.02$ & 0.895 & 0.868 \\
 $\textit{TIFF(ours)}$ & $\textbf{0.828}\pm 0.02$&\textbf{0.919}& \textbf{0.906}\\ 
 \hline
\end{tabular}
\label{Table4}
\end{table}

We can clearly found that focusing only on one type of local information does not produce the best results, but it is still significantly better than focusing only on Global Information. Adding Patch-based local information to Global Information will increase the AUC to 0.810, and adding Feature-based local information to Global Information will increase the AUC to 0.806. However, Tri level Information Fusion mechanism combining the three information achieved the best result with an AUC of 0.828.

\subsection{Implementation Details}
In this study, we evaluated the breast cancer early screening task on two public datasets using various approaches, including MIL, Single-view, and Multi-view methods, and compared them with our proposed Tri-level Information Fusion Context Clustering Framework. All experiments were conducted on a single NVIDIA 3090 24G GPU, using Adam as the optimizer \cite{80}. A fixed-step learning rate (StepLR) decay strategy was employed to fine-tune the learning rate, preventing overfitting and ensuring better convergence to the optimal solution.

\section{Discussion}

Although our model achieves state-of-the-art performance on two public datasets, the varying performance of MIL across different datasets has drawn our attention. We believe the poor performance of MIL on the Vindr-Mammo dataset is due to its long-tail distribution. Addressing the impact of long-tail distribution and exploring MIL’s broader performance on well-balanced datasets will be an interesting research direction in the future.

Exploring ways to better assist clinicians in screening tasks to reduce their workload also represents a more promising research avenue.

\section{Acknowledgments}
The author would like to acknowledge the funding support from the Clinical Research Program of the First Affiliated Hospital of Shenzhen University (2023YJLCYJ019) for this work.

\section{Conclusions}
This study presents a novel, weakly supervised multi-view tri-level information fusion framework for early breast cancer screening using mammography images. Unlike conventional feature extraction paradigms such as CNN and ViT, our approach leverages a context-clustering-based methodology. This paradigm enhances computational efficiency and facilitates easier association of structural or pathological features, making it suitable for clinical tasks in mammography. Additionally, through the tri-level information fusion framework, the proposed model effectively integrates complementary information from different levels, significantly enhancing diagnostic accuracy.
Comprehensive evaluations conducted on two publicly available datasets, Vindr-Mammo and CBIS-DDSM, showcase the model's exceptional performance, achieving state-of-the-art accuracy. These results underscore the potential of the proposed framework as a robust and scalable solution for early breast cancer detection, offering significant promise for deployment in clinical settings.

\bibliographystyle{ieeetr}
\bibliography{ref}

\end{document}